\pdfoutput=1

\documentclass[11pt]{article}

\usepackage[final]{acl}
\usepackage[utf8]{inputenc}
\usepackage{microtype}
\usepackage{times}

\usepackage{enumitem}

\usepackage{wrapfig}
\usepackage{subcaption}
\usepackage{makecell}
\usepackage{siunitx}
\usepackage{amsmath}
\usepackage{graphicx}
\usepackage{graphbox}
\usepackage{float}
\usepackage{caption}

\title{Understanding Players as if They Are Talking to the Game in a Customized Language: A Pilot Study}

\author{
\textbf{Tianze Wang\textsuperscript{1,2}\thanks{Both authors contributed equally to this research.}}
\textbf{Maryam Honari-Jahromi\textsuperscript{2}\footnotemark[1]}
\\
\textbf{Styliani Katsarou\textsuperscript{2}}
\textbf{Olga Mikheeva\textsuperscript{1,2}}
\textbf{Theodoros Panagiotakopoulos\textsuperscript{2}}
\\
\textbf{Oleg Smirnov\textsuperscript{2}\thanks{Joint senior authorship.
}\thanks{Corresponding author:~\href{mailto:oleg.smirnov@microsoft.com}{oleg.smirnov@microsoft.com}
}}
\textbf{Lele Cao\textsuperscript{2}\footnotemark[2]}
\textbf{Sahar Asadi\textsuperscript{2}\footnotemark[2]}
\\
  \textsuperscript{1}KTH Royal Institute of Technology
  \textsuperscript{2}King, Microsoft Gaming
}

\begin{document}
\maketitle

\begin{abstract}
This pilot study explores the application of language models (LMs) to model game event sequences, treating them as a customized language. We investigate a popular mobile game, transforming raw event data into textual sequences and pretraining a Longformer model on this data. Our approach captures the rich and nuanced interactions within game sessions, effectively identifying meaningful player segments. The results demonstrate the potential of self-supervised LMs in enhancing game design and personalization without relying on ground-truth labels.
\end{abstract}

\section{Introduction}
\label{sec:introduction}

The dominant form of human interaction is natural language, represented by a {\it stream of words}. Language Models (LMs) have become highly effective in understanding and representing these general-purpose natural languages. Similarly, when a human player interacts with a video game, the primary form of interaction is through game controls, which lead to visual and auditory feedback. This in-game interaction is typically recorded as a {\it stream of events}, each with rich attributes and categories.
This pilot study explores {\bf whether we can apply LMs, initially designed for word sequences, to model game event sequences}. 
Understanding player behavior through this modeling approach is crucial for designing engaging experiences, improving game mechanics, and personalizing content. For example, understanding the optimal balance between challenge and progression can enable dynamic game difficulty adjustments, maximizing the enjoyment experienced by players.

Traditionally, understanding game players has relied on surveys and interviews, such as those conducted in \cite{rodrigues2022automating}. While these methods provide valuable insights, they are significantly limited by scalability. Deep Learning (DL) models, like those in \cite{cao2020debiasing}, have been trained on aggregated (from game events) gameplay data to achieve in-game personalization, but they often neglect nuanced interactions. Recently, training DL models on sequential interactions between players and in-game items has been explored, as exemplified by \cite{villa2020interpretable}.
However, these modeled interactions are still relatively limited in type and richness compared to game events.
Moreover, most of these DL models only optimize for specific personalization scenarios, requiring large amount of ground-truth labels, which are not always available.

As a consequence, self-supervised LM pretraining emerged as a promising approach to directly model the rich and fine-grained game events in a scalable way without requiring any labels. In principle, this pretrained model is not restricted to any specific personalization use case. 
To the best of our knowledge, this is the first attempt to pretrain an LM on game events by treating these events as a customized natural language.
The highlights of this pilot study are: (\S\ref{sec:game_events}) studying a popular mobile video game from King\footnote{https://king.com}, Candy Crush Saga, (\S\ref{sec:events2words}) developing a simple method for transforming a large amount of game events into language tokens, (\S\ref{sec:pretrain_lm}) pretraining an LM on the customized ``language'' representing game events, (\S\ref{sec:results}) reporting experimental results on the LM’s intrinsic performance and its capability in understanding game players, and finally (\S\ref{sec:ethical_consideration}) we outline measures employed to mitigate ethical considerations.

\begin{figure*}[t!]
    \begin{center}
        \includegraphics[width=\textwidth]{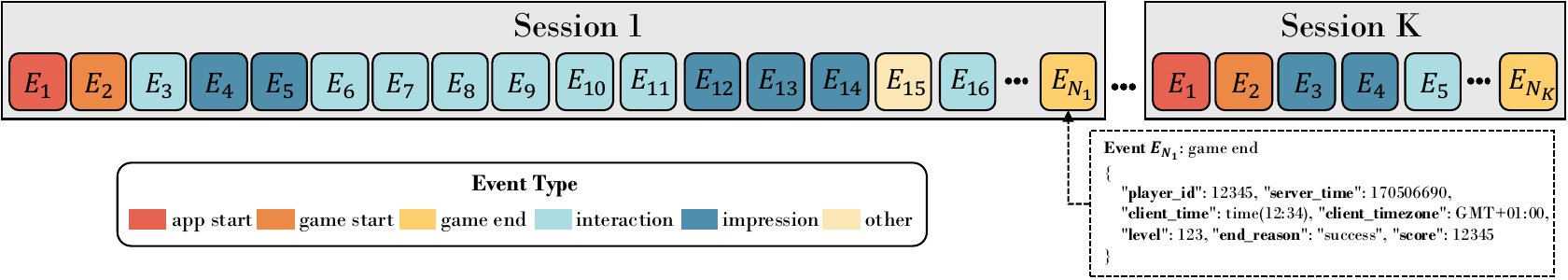}
        \caption{Example events segmented into semantic sessions. The final game-end event in ``Session 1'' is expanded to show details about its associated fields and values.}
        \label{fig:sessions}
    \end{center}
\end{figure*}

\section{Related Work}
\label{sec:related_work}

Modeling sequential interactions between users and items has been extensively studied in recommendation systems. Initial approaches utilized Markovian assumptions for collaborative filtering~\cite{zimdars2001using}, later extended to Markov decision processes~\cite{shani2005mdp}. Predicting future behavior trajectories using contextual and sequential information has been addressed with autoregressive Long Short-Term Memory models~\cite{wu2017recurrent} and coupled Recurrent Neural Network (RNN) architectures for joint modeling of user/item interactions~\cite{kumar2019predicting}. Explicitly modeling different types of user behavior, such as repeated consumption, has also shown to improve downstream performance metrics~\cite{anderson2014dynamics, ren2019repeatnet}.

LMs have been leveraged for embedding sequential data in recommendation settings, beginning with music track representations using the Word2Vec objective~\cite{mehrotra2018towards} and extending to modeling sequences of listening sessions with RNNs~\cite{hansen2020contextual}. More recently, self-attention sequential models have been introduced, such as BERT4Rec~\cite{sun2019bert4rec}, which balance the trade-off between Markov chain models and neural network methods. Follow-up work on multi-task customer models for personalization has further advanced this field by integrating novel data augmentation and task-aware readout modules~\cite{luo2023mcm}.

Despite these advancements, the application of LMs for user modeling in gaming remains underexplored. Our study proposes the first approach for learning representations of mobile game players by pretraining a Transformer architecture in a self-supervised manner, treating game event sequences as a customized natural language. This approach aims to capture the rich and nuanced interactions within game sessions.

\section{The Game and Interaction Events}
\label{sec:game_events}

This pilot study focuses on Candy Crush Saga game. When a player interacts with this game on a mobile device, their behavior generates a sequence of time-ordered events, which are recorded locally on the user’s device and later sent to the central game server in batches. Example events include starting the game application, beginning a new game round, purchasing in-game items, and displaying pop-ups and notifications. The tracked player behavior events fall into 12 categories, each with an associated schema containing continuous and categorical features.

\begin{figure}[t!]
    \centering
    \begin{subfigure}[t]{0.49\columnwidth}
        \includegraphics[trim=0.2cm 0.2cm 0.7cm 0.2cm, clip,align=c,height=4cm]{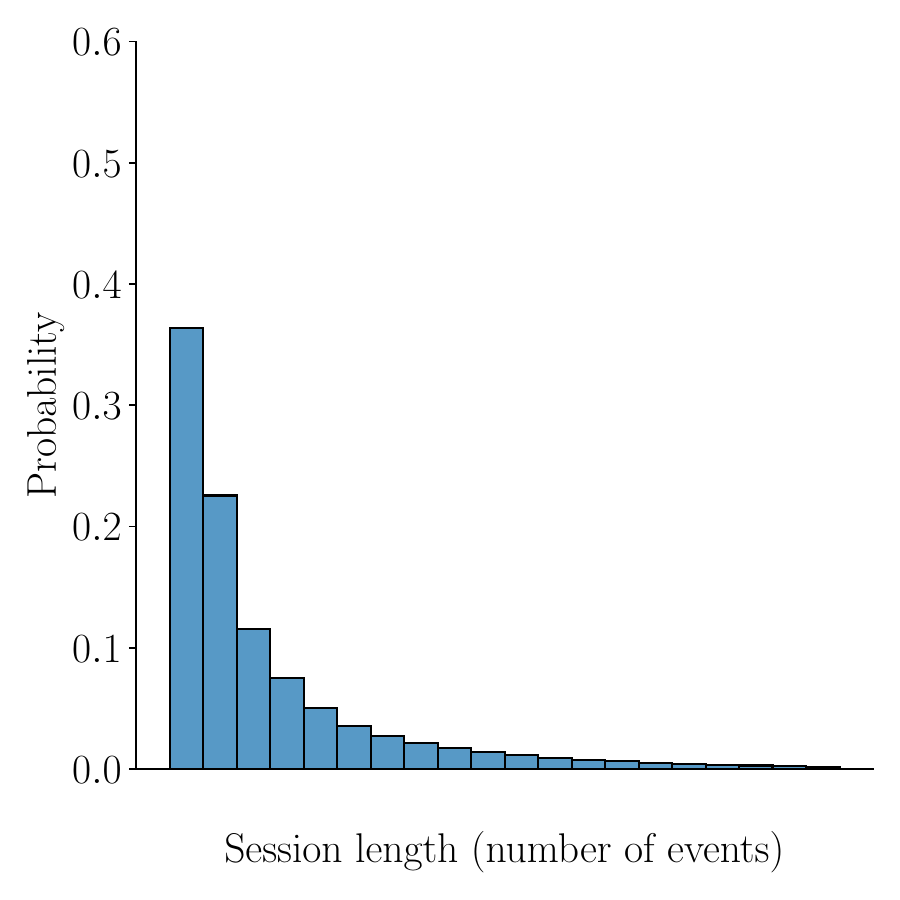}
        \caption{}
        \label{fig:dataset_session_length}
    \end{subfigure}
    \begin{subfigure}[t]{0.49\columnwidth}
        \includegraphics[trim=0.2cm 0.2cm 0.7cm 0.2cm, clip,align=c,height=4cm]{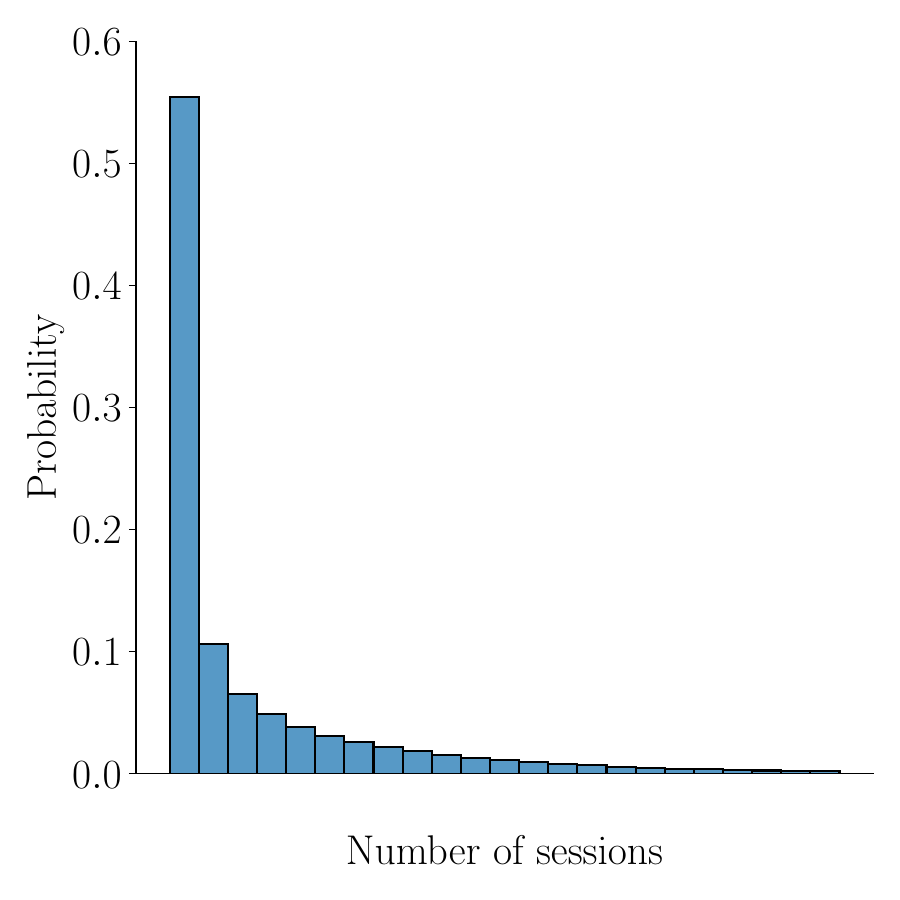}
        \caption{}
        \label{fig:dataset_activity_dist}
    \end{subfigure}
    \caption{(a) Histogram of session lengths and (b) the distribution of session quantities over a 15-day period shown up to the 99\textsuperscript{th} percentile.}
\end{figure}

The player-game interaction events are segmented into sessions based on the player's activity semantics, as illustrated in Figure~\ref{fig:sessions}. According to game analytics conventions recommended by the data scientists from the game producer, a session is considered to have ended if a player is inactive for 15 minutes or more. For this study, we collected a dataset of player event sessions over 15 days, with $10{,}000$ players uniformly sampled from the entire player population. The resulting dataset consists of $125{,}000$ sessions, split into a 2:1 train-test ratio. The distribution of session lengths in the dataset is shown in Figure~\ref{fig:dataset_session_length}, while Figure~\ref{fig:dataset_activity_dist} depicts the distribution of sessions quantities. Both session lengths and quantities approximately follow a geometric distribution.

Our collected event data, while superficially similar to tracking data in other domains like e-commerce, presents unique challenges. In-game interactions occur at a much higher frequency than in web browsing, resulting in large volumes of potentially redundant events that call for careful preprocessing and modeling of long-range dependencies. Additionally, game event sequences are often noisy, with incorrectly ordered events or missing ordering information due to users switching between online and offline modes, which can degrade model performance during training and inference.

\begin{figure*}[t!]
    \centering
    \includegraphics[width=\textwidth]{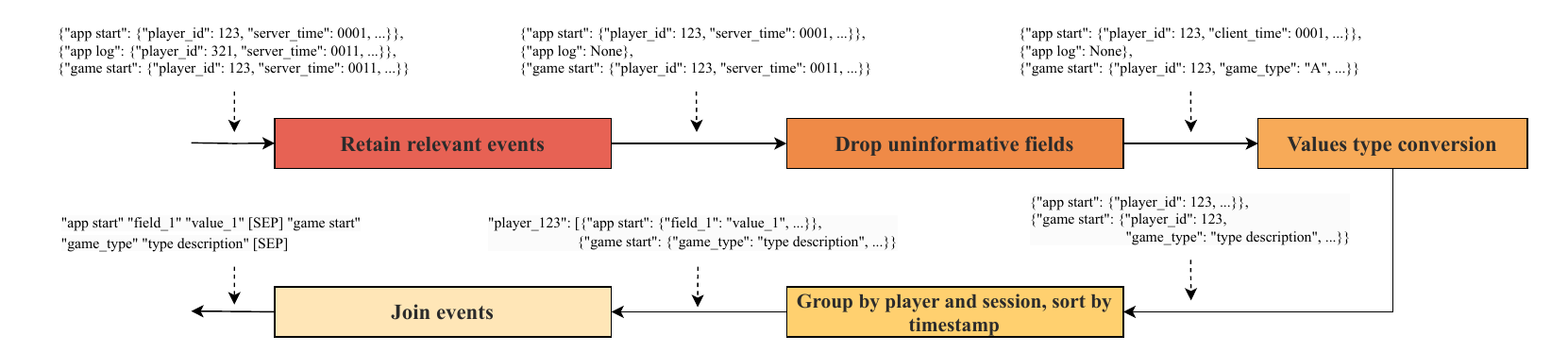}
    \caption{The pipeline to convert event streams to word streams.}
    \label{fig:preprocessing_diagram}
\end{figure*}

\section{From Events to Words}
\label{sec:events2words}

The raw format of game events is JSON. To make this data digestible by LMs, we designed a simple pipeline to transform raw events into textual sequences. As illustrated in Figure~\ref{fig:preprocessing_diagram}, the pipeline begins by removing unnecessary events and fields. Leveraging game-specific knowledge, we filter out non-informative data, such as device-specific logs, reducing the number of event fields by over 90\%. We bin certain numerical features, such as the hour of the day, based on domain-specific knowledge to convert them into categorical variables. Additionally, we group similar in-game event identifiers, e.g., the name of the UI shown, to reduce the vocabulary size.
The words are then grouped by users and sessions, ordered by timestamps to preserve the natural interaction flow, and concatenated to form a textual description of a player’s interaction experience.

We use a word-level tokenizer that splits a space-separated string into tokens and maps them to unique identifiers. This approach suits the relatively small vocabulary of behavior data (${\sim}13{,}500$ tokens), though the tokenized sequences are much longer than those in typical NLP tasks like sentiment analysis.

\section{Pretrain a Language Model}
\label{sec:pretrain_lm}

The tokenized word sequences are often longer than 512 tokens, which are unmanageable for the conventional BERT~\cite{kenton2019bert} architecture and its derivatives. Modeling long sequences poses a significant challenge to Transformer-based approaches due to the self-attention operation, which scales quadratically with input length in terms of memory and computational complexity. This challenge is intensified when modeling distant dependencies in extended gameplay experiences that involve concatenating multiple sessions. To overcome this, we adopt Longformer~\cite{beltagy2020longformer}, a model designed specifically for processing long textual inputs.

Longformer combines dilated sliding window attention for local context and global attention on a few pre-selected input locations. This approach scales linearly with input size, enabling the processing of sequences up to $4{,}096$ tokens in a single pass, which is sufficient for most behavior modeling scenarios. Additionally, Longformer’s sparse attention pattern performs well in contexts where many tokens in the immediate local context may be redundant, as is often the case with high-frequency game events.

We pretrained several Longformer variants\footnote{We use the HuggingFace Transformers~\cite{wolf2019huggingface}  library and PyTorch framework~\cite{paszke2019pytorch} for model implementation. All models were trained with half-precision (FP16) on a single NVIDIA A100 GPU, with the \emph{large} model taking approximately 50 hours to complete pretraining.} from scratch with different capacities, based on the hyper-parameters listed in Table~\ref{tbl:hyperparams}. We experimented with the baseline Longformer configuration, i.e., ``\emph{large}’’, and two smaller model variants with fewer internal layers and self-attention heads. The models were optimized with the masked language modeling (MLM) objective using Adam~\cite{kingma2014adam} with a fixed learning rate of $2 \times 10^{-5}$. Each LM was trained from randomly initialized weights for 100 epochs with a batch size of 4 and gradient accumulation over 4 steps, resulting in an effective batch size of 16 ($2^{16}$ tokens).

\begin{table}[t!]
\begin{center}
\small
\addtolength{\tabcolsep}{-3.2pt}
\begin{tabular}{rrrrrr} 
 \hline
 model size & \#layer & \#head & dims & block size & \#params \\ 
 \hline
 \emph{small} & 2 & 2 & 128 & 1024& 2M \\ 
 \emph{medium} & 6 & 6 & 384 & 2048 & 20M \\ 
 \emph{large} & 12 & 12 & 768 & 4096 & 121M \\ 
 \hline
\end{tabular}
\caption{Hyperparameters for different model sizes.}
\label{tbl:hyperparams}
\end{center}
\end{table}

\begin{table}[t!]
\begin{center}
\small
\addtolength{\tabcolsep}{-2.2pt}
\renewcommand{\arraystretch}{1.2}
\begin{tabular}{rrrrr} 
\hline

 model size & accuracy~$\uparrow$ & perplexity~$\downarrow$ & CE~$\downarrow$ \\ [0.5ex] 
\hline
 \emph{small} &  $ 0.69 \pm 0.06 $ & $ 3.27 \pm 0.71 $ &  $ 1.16 \pm 0.22$ \\ 
 \emph{medium} & $ 0.93 \pm 0.01 $ & $ 1.28 \pm 0.09 $ & $ 0.25 \pm 0.07$  \\ 
 \emph{large} &  $ \mathbf{0.95} \pm \mathbf{0.01} $ & $ \mathbf{1.16} \pm \mathbf{0.05} $ & $\mathbf{0.15}\pm \mathbf{0.04}$ \\ 
\hline
 
\end{tabular}
\caption{Mean values and standard deviations of intrinsic language modeling metrics computed over five training runs.}
\label{tbl:metrics}
\end{center}
\end{table}

\section{Results}
\label{sec:results}

First, we evaluate the intrinsic performance of the proposed approach using intrinsic MLM metrics. We report the Cross-Entropy (CE) loss and multi-class classification accuracy of predicting masked tokens on the validation split for the tested model architectures, as shown in Table~\ref{tbl:metrics}. Additionally, we report the perplexity score, following established methodologies for evaluating MLM pretraining performance~\cite{liu2019roberta}. As expected, we observe that LMs with larger capacities achieve better fits for the behavior sessions without overfitting.

\begin{figure}[t!]
    \centering
     \begin{subfigure}[t]{0.46\textwidth} 
        \includegraphics
        [trim=0.2cm 1cm 0cm 0.2cm, clip, align=l, width=\textwidth]
        {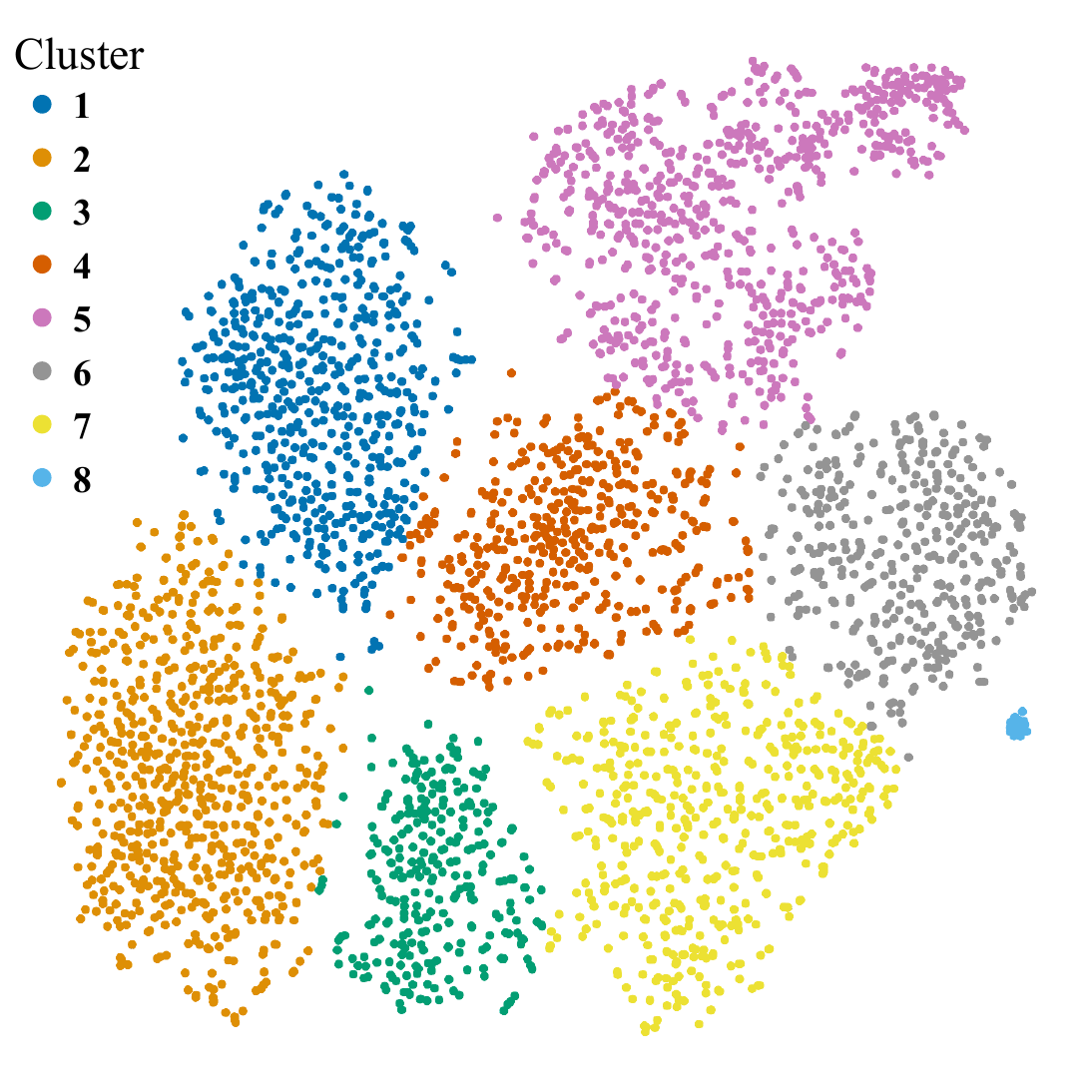}
        \caption{} \label{fig:clustering-tsne}
    \end{subfigure}
    \\
    \vspace{10pt}
     \begin{subfigure}[t]{0.46\textwidth} 
        \includegraphics[trim=-1cm 0.1cm 0cm 0cm, clip, align=r, width=\textwidth]{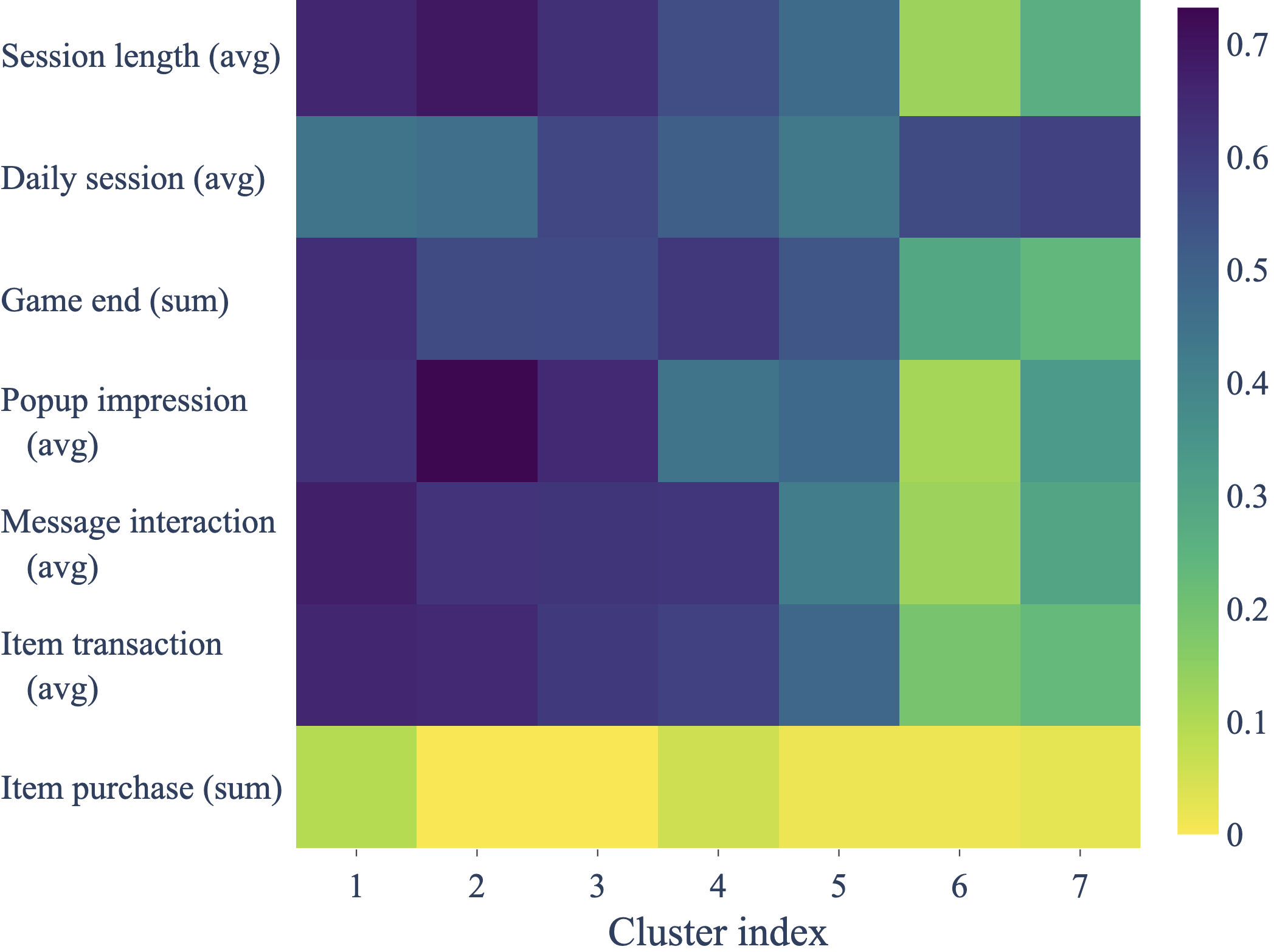}
        \caption{} \label{fig:clustering-fingerprint}
    \end{subfigure}
    ~
    \caption{(a) t-SNE of the latent embedding space from the pretrained \emph{large} Longformer with Gaussian Mixture Model clustering. (b) Histogram of quantized player events in clusters (excluding cluster 8 due to small size and lack of gameplay).
    }
    \label{fig:tsne_plot}
\end{figure}

Next, we perform a qualitative analysis to identify spontaneous player clusters representing different behavioral persona. We extract embeddings of input token sequences from the pretrained \emph{large} Longformer model. Using $4096\times768$-dim representations from the last Attention layer, we apply max pooling over sequence length to compute an embedding vector for each input sequence. These session embeddings are projected onto the first 50 principal components using linear PCA to reduce noise and speed up computation. The projections are then mapped to 2D space via t-SNE~\cite{van2008visualizing} and clustered with a Gaussian Mixture Model~\cite{gmmreynolds2009} with eight components. The resulting t-SNE plot is shown in Figure~\ref{fig:clustering-tsne}. Analyzing the average player behavior within the well-separated t-SNE clusters in Figure~\ref{fig:clustering-fingerprint}, we collaboratively identified player segments with game analysts from a practical product perspective. Identified players' personas qualitatively resonate with what our user researchers extracted from self-reported behavioral surveys:
\begin{enumerate}[leftmargin=*, itemsep=0pt]
\item {\it Competitive devoted}: a skilled player who plays less often but long sessions, occasionally purchasing items and collecting utilities.
\item {\it Casual devoted}: a player who plays long sessions infrequently, engages in quests, collects rewards, and prefers free gameplay.
\item {\it Persistent devoted}: a player who plays frequent, long sessions without purchasing.
\item {\it Lean-in casual economy aware}: A skilled player who plays less often but for long sessions, occasionally buying items.
\item {\it Lean-in casual}: a skilled player who plays less often but for long sessions.
\item {\it Persistent casual}: a less skillful player who plays short, frequent sessions with little engagement in social and economic aspects.
\item {\it Persistent collector}: a player with frequent short sessions, collecting utilities to progress.
\end{enumerate}


\section{Ethical Considerations}
\label{sec:ethical_consideration}

Computational modeling of player behavior in games has raised various ethical concerns within both research and industry~\cite{mikkelsen2017ethical}. In this pilot study, we utilize non-personally identifiable tracking data from in-game interactions to create vectorized representations of player behaviors. Our objective is to leverage these representations to support personalized and enhanced player experiences while maintaining ethical standards.

Potential ethical risks include (1) biases in the input dataset, such as under-representing less frequent player behaviors, and (2) the misapplication of models to different data distributions, known as Type III errors~\cite{mikkelsen2017ethical}. To mitigate these risks, we use robust data validation and automated model analysis tools available in production-ready machine learning frameworks~\cite{46484}.

We address under-represented player behaviors through qualitative evaluation methods, such as embedding space visualization. Additionally, we periodically retrain the model with recent data to address distribution shifts, with retraining intervals determined empirically based on model performance and data drift.

For the downstream recommendation system, we plan to implement model explainability and uncertainty estimation methods to better understand the model’s robustness, biases, and other ethical considerations. These measures aim to ensure that our modeling approach supports ethical and responsible AI deployment.

\section{Conclusion and Future Work}
\label{sec:conclusion}
This pilot study demonstrates the potential of using self-supervised language models to understand player behavior by modeling game event sequences as a customized natural language. Our approach, leverages the Longformer model to effectively captures the rich and nuanced interactions within game sessions in a self-supervised manner, agnostic to downstream use-cases. The results highlight the model’s ability to identify meaningful player segments, providing valuable insights for game design and personalization. For future work, we plan to extend training to single- and multitask fine-tuning with labeled datasets to benchmark against fully-supervised baselines. We anticipate that our approach can be extended to other event-based game datasets as well.

\bibliography{player2vec_emnlp}

\end{document}